\pdfoutput=1

\documentclass[11pt]{article}

\usepackage[preprint]{acl}

\usepackage{times}
\usepackage{latexsym}

\usepackage[T1]{fontenc}

\usepackage[utf8]{inputenc}

\usepackage{microtype}

\usepackage{inconsolata}

\usepackage{graphicx}

\usepackage{booktabs}

\title{Affective Polarization across European Parliaments}

\author{
 \textbf{Bojan Evkoski~\textsuperscript{1}},
 \textbf{Igor Mozetič~\textsuperscript{2}},
 \textbf{Nikola Ljubešić~\textsuperscript{2, 3, 4}},
 \textbf{Petra Kralj Novak~\textsuperscript{1, 2}}
 \\\\
 \textsuperscript{1} Central European University, Vienna, Austria
 \\
 \textsuperscript{2} Jožef Stefan Institute, Ljubljana, Slovenia
 \\
 \textsuperscript{3} University of Ljubljana, Ljubljana, Slovenia
 \\
 \textsuperscript{4} Institute of Contemporary History, Ljubljana, Slovenia
\\\\
 \small{
   \textbf{Correspondence:} \href{mailto:evkoski_bojan@phd.ceu.edu}{evkoski\_bojan@phd.ceu.edu}
 }
}

\begin{document}
\maketitle
\begin{abstract}
Affective polarization, characterized by increased negativity and hostility towards opposing groups, has become a prominent feature of political discourse worldwide. Our study examines the presence of this type of polarization in a selection of European parliaments in a fully automated manner. Utilizing a comprehensive corpus of parliamentary speeches from the parliaments of six European countries, we employ natural language processing techniques to estimate parliamentarian sentiment. By comparing the levels of negativity conveyed in references to individuals from opposing groups versus one's own, we discover patterns of affectively polarized interactions. The findings demonstrate the existence of consistent affective polarization across all six European parliaments. Although activity correlates with negativity, there is no observed difference in affective polarization between less active and more active members of parliament. Finally, we show that reciprocity is a contributing mechanism in affective polarization between parliamentarians across all six parliaments.
\end{abstract}

\section{Introduction}

Political polarization, the widening gap between differing ideological groups, impacts governance, public discourse, and social unity. This division, powered by media echo chambers, confirmation biases, and identity politics, requires empathetic communication and evidence-based discussions to promote a more harmonious political landscape.

Within political science literature, two primary categories of polarization mechanisms are widely recognized: \textit{ideological} and \textit{affective polarization }\cite{hohmann2023quantifying, kubin2021role, iyengar2019origins}. 

\textit{Ideological polarization} refers to the divergence of ideologies and a decline in dialogue among individuals holding different views. 
\textit{Affective polarization} referes to the extent to which people have affinity towards their political allies (in-group members) and hostility towards their political 
opponents (out-group members) \cite{iyengar2012affect}. Although these two forms of polarization can reinforce each other, they are distinct concepts both in terms of theoretical underpinnings and empirical measurements \cite{dias2022nature}. While measuring ideological polarization relies on data about people's opinions on a certain topic, assessing affective polarization requires information about the emotional dynamics between groups \cite{druckman2019we}. 

Traditional studies on affective polarization predominantly rely on public opinion polls to measure individuals' negative feelings towards opposing groups \cite{bettarelli2023regional, kekkonen2021affective, hobolt2021divided}. Contemporary approaches use automatic methods such as sentiment analysis to study affectiveness between structurally diverging groups on social platforms \cite{tyagi2020affective, lerman2024affective}. However, past research has typically focused on the general public rather than the behaviors and rhetoric of public political figures who play a crucial role in creating the public discourse itself \cite{matsubayashi2013politicians}.

Healthy democracies thrive on the competition and negotiation between opposing sides inside institutions such as parliaments. Political scientists emphasize that even contentious debates are preferable to the absence of dialogue, as frequent interactions between diverse viewpoints can help prevent the entrenchment of extreme polarization \cite{harris1998democracy}. Therefore, it is essential to study affective polarization among politicians within political elites. By doing so, we can understand the dynamics of political discourse and address affective polarization before it becomes so entrenched that meaningful dialogue is no longer possible.

With the emergence of big data collections from parliaments in recent years \cite{mollin2007hansard, erjavec2023parlamint, european_parliament_meetings_2024} combined with the sophistication of automatic text processing tools, we have the opportunity to explore affective polarization of politicians on a new scale. In this work, we conduct a fully data-driven quantitative assessment and analysis of affective polarization of members of parliaments (MPs) within six national European parliaments. We employ a methodology that combines an LLM-based sentiment model fine-tuned on parliamentary speeches with named entity recognition and disambiguation to assess affectiveness of MPs towards one another. The goal of the approach is to estimate if negativity of speeches is higher when directed toward members of opposing groups. 

We present three main findings. First, affective polarization does occur between Coalition and Opposition in all six studied European parliaments. Second, the more active parliamentarians exhibit higher levels of negativity with no observed difference in their affective polarization. Third, we show that reciprocity is a contributing mechanism in the affective polarization between MPs.

\section{Data}
We analyze transcriptions of speeches from parliaments of six European countries which exemplify all major regions of Europe: Denmark (Folketing, Northern Europe), France (Assemblée nationale, Western Europe), Poland (Sejm, Central Europe), Serbia (Narodna Skupština, South-Eastern Europe), Spain (Congreso de los Diputados, South-Western Europe) and Ukraine (Verkhovna Rada, Eastern Europe).
The transcriptions we use are part of the ParlaMint 4.0 dataset \cite{parlamint4, erjavec2023parlamint}, which is a collection of 29 multilingual corpora consisting of parliamentary debates from 2015 to mid-2022 (with several exceptions). The corpora are between 9 and 125 million words in size and contain extensive metadata, including information about the parliament, speakers, and speeches. The dataset also includes marked-up transcriber comments and additional information such as the speakers' year of birth and links to their Wikipedia pages. Linguistic annotations, such as tokenization, sentence segmentation, part-of-speech tagging, and syntactic dependencies are provided, along with named entity annotations. Table~\ref{tab:parliament_stats} shows the general statistics of the ParlaMint data of the six selected parliaments.

\begin{table*}[htp!]
\centering
\footnotesize
\caption{General statistics of the analyzed data for the six selected parliaments.
Speeches include all speeches with at least five sentences held by regular MPs (Members of Parliament), 
excluding chairpersons or guests.
Abbreviations: Coa (coalition), Opp (opposition), WPS (word per speech).
}

\label{tab:parliament_stats}
\begin{tabular}{r c c c c c c} 
\toprule
Parliament & Terms & From & To & Speeches (Coa/Opp) & WPS & MPs (Coa/Opp) \\ 
\hline
Denmark (DK) & 3 & 2014-10-07 & 2022-06-07 & 127K (39K/59K) & 217.77 & 368 (158/239) \\
France (FR) & 2 & 2017-06-27 & 2022-06-29 & 57K (22K/25K) & 215.79 & 455 (299/131)\\
Poland (PL) & 2 & 2015-11-12 & 2022-06-23 & 59K (18K/35K) & 258.93 & 623 (250/289) \\
Serbia (RS) & 9 & 1997-12-03 & 2022-02-14 & 120K (70K/47K) & 524.33 & 1230 (861/420)\\
Spain (ES) & 5 & 2015-01-20 & 2020-12-15 & 17K (5K/11K) & 695.67 & 715 (308/497)\\
Ukraine (UA) & 3 & 2012-12-04 & 2023-02-24 & 60K (23K/25K) & 167.64 & 936 (683/282) \\ 
\bottomrule
\end{tabular}
\vspace{-1em}
\end{table*}

For this study, we automatically label each speech with a sentiment score using the ParlaSent multilingual large language model (LLM), an XLM-R RoBERTa model fine-tuned for sentiment analysis on parliamentary speeches in five languages \cite{mochtak2023parlasent}. Before applying the model, we perform minimal preprocessing by excluding the first and last sentences of each speech, as they are typically procedural and positive but lack significant content. The output of the model is a continuous sentiment score from 0 (most negative) to 5 (most positive).

\section{Results}

\subsection{Sentiment Distribution}

\begin{figure}[t]
    \centering
    \includegraphics[width=0.99\linewidth]{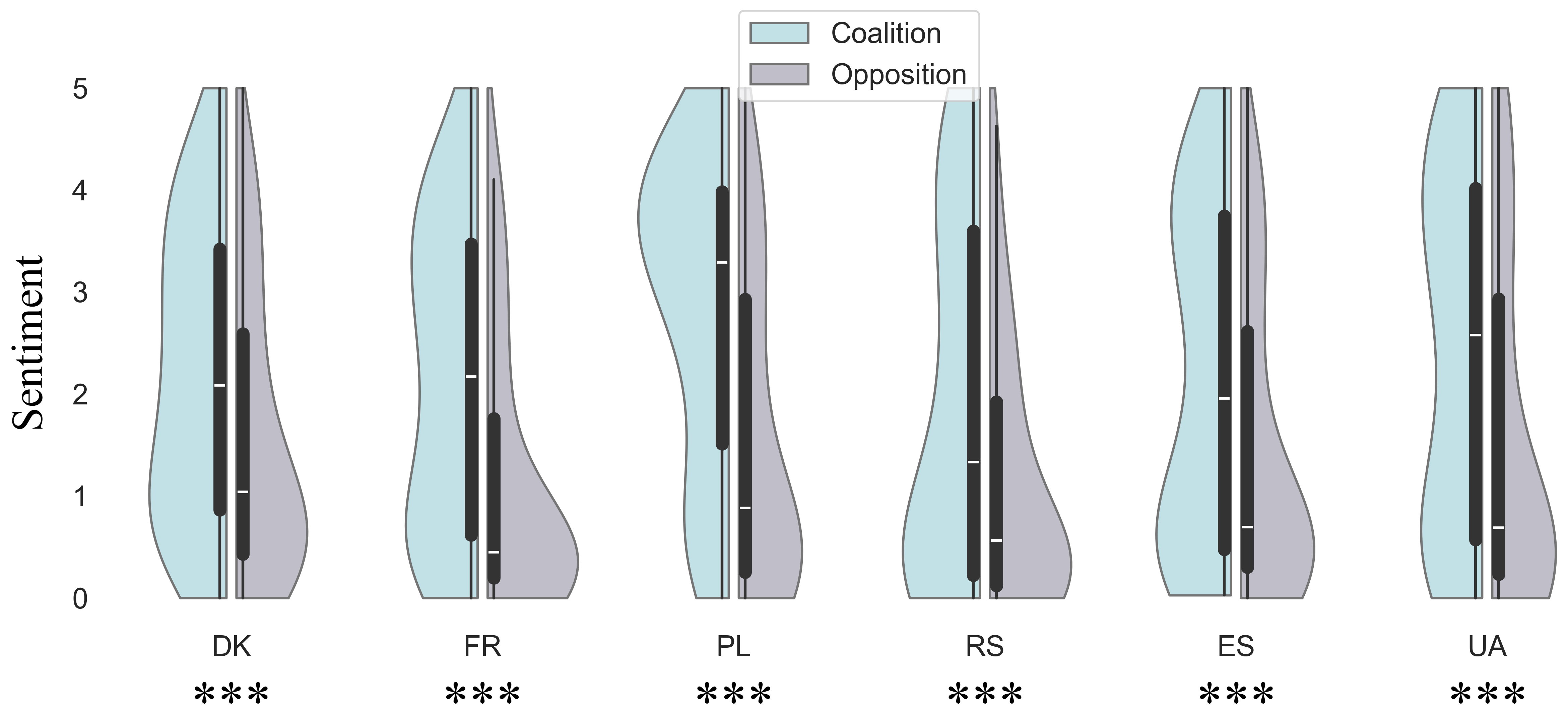}
    \caption{Sentiment distribution of the Coalition and Opposition in six selected European parliaments. Asterisks represent the statistical significance in the comparison of the distributions using a Kolmogorov-Smirnov test.}
    \vspace{-1em}
    \label{fig:fig1}
\end{figure}

Before presenting the assessment of affective polarization, we first show the sentiment distribution in parliamentary speeches delivered by both Coalition and Opposition MPs. Notably, there is a prevailing negative sentiment inclination in all six parliaments, with the median score falling below the neutral value of 2.5. Our results also show that in all six parliaments, the Opposition is significantly more negative than the Coalition (see Figure \ref{fig:fig1}). The largest difference between Opposition and Coalition negativity is in the parliaments of France and Poland.

Acknowledging the individuality of parliamentary customs, the diverse temporal scopes considered for each parliament, as well as the variability in the performance of language tools, we refrain from conducting direct cross-parliament comparisons of sentiment distributions. Instead, we invite the reader to observe each parliament individually, emphasizing the distinctions between Coalition and Opposition groups and their influence on affective polarization within each legislative body.

\subsection{Affective Polarization}

In exploring affective polarization, our focus is on quantifying the disparity between sentiments expressed in references to in-group and out-group MPs. To achieve this, we use the Named Entity tags embedded in the ParlaMint dataset to identify in-group and out-group entity references, with the inclusion of several steps of Named Entity disambiguation (see the Appendix for details). We then calculate the sentiment scores of speeches directed towards the same group (e.g., a Coalition MP referring to a Coalition MP) and towards the other group, respectively. We conduct separate assessments for the Coalition and the Opposition.

\begin{figure}
    \centering
    \includegraphics[width=1\linewidth]{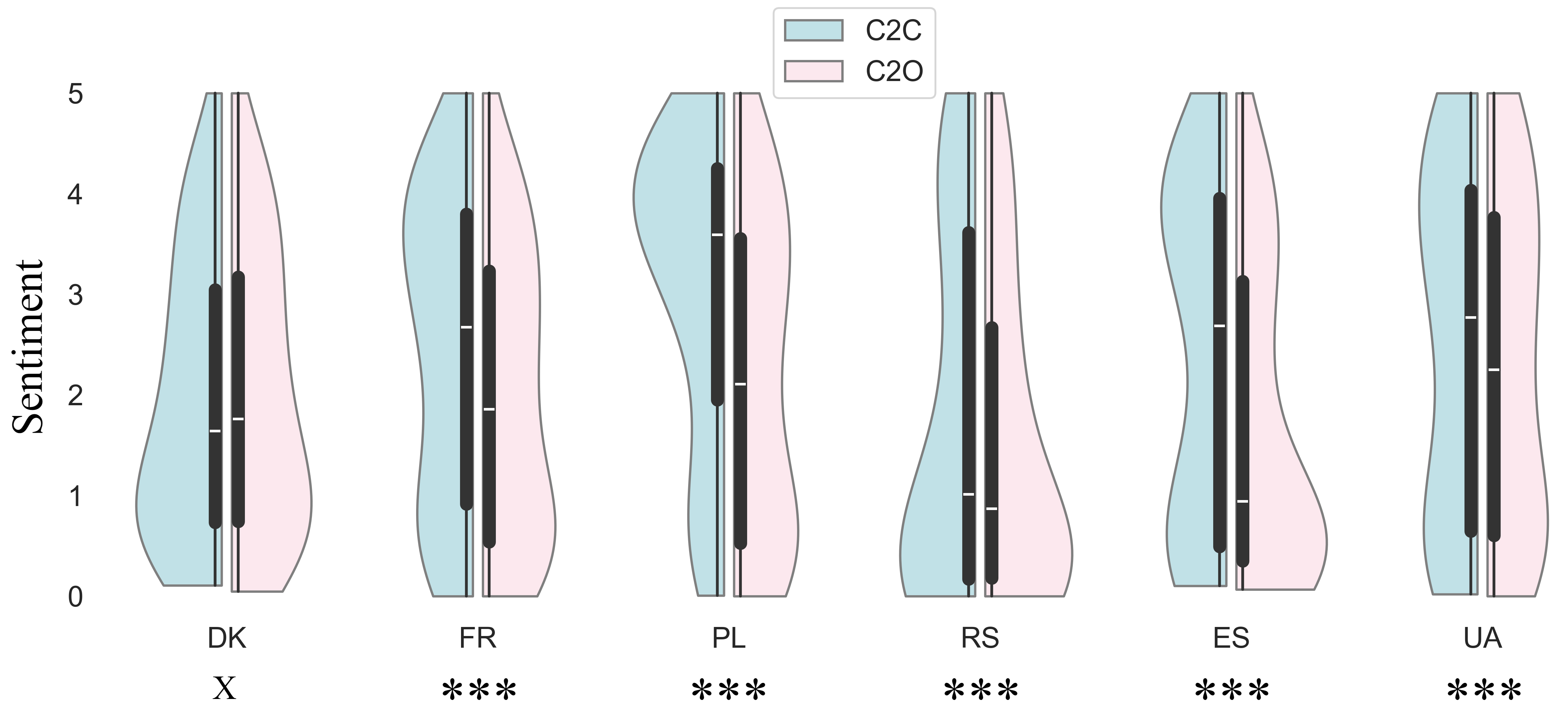}
    \caption{Sentiment distribution of the Coalition towards Coalition (C2C) and towards Opposition (C2O).}
    \vspace{-1em}
    \label{fig:fig2}
\end{figure}

The outcomes for Coalitions are depicted in Figure \ref{fig:fig2}, illustrating the comparison between the C2C sentiment distributions (a Coalition MP referring to a fellow Coalition MP) and the C2O sentiment distributions (a Coalition MP referring to an Opposition MP). 
The results reveal that across five out of the six analyzed parliaments (Denmark being the exception), Coalitions exhibit greater negativity when referencing their Opposition.  
This trend is especially evident in France, Poland, and Spain, where intra-coalition discourse remains positive, contrasting sharply with the negative tone when mentioning their Opposition. 
These findings provide evidence of political divisions characterized by dislike, which underscores the polarized nature of political communication.

\begin{figure}
    \centering
    \includegraphics[width=1\linewidth]{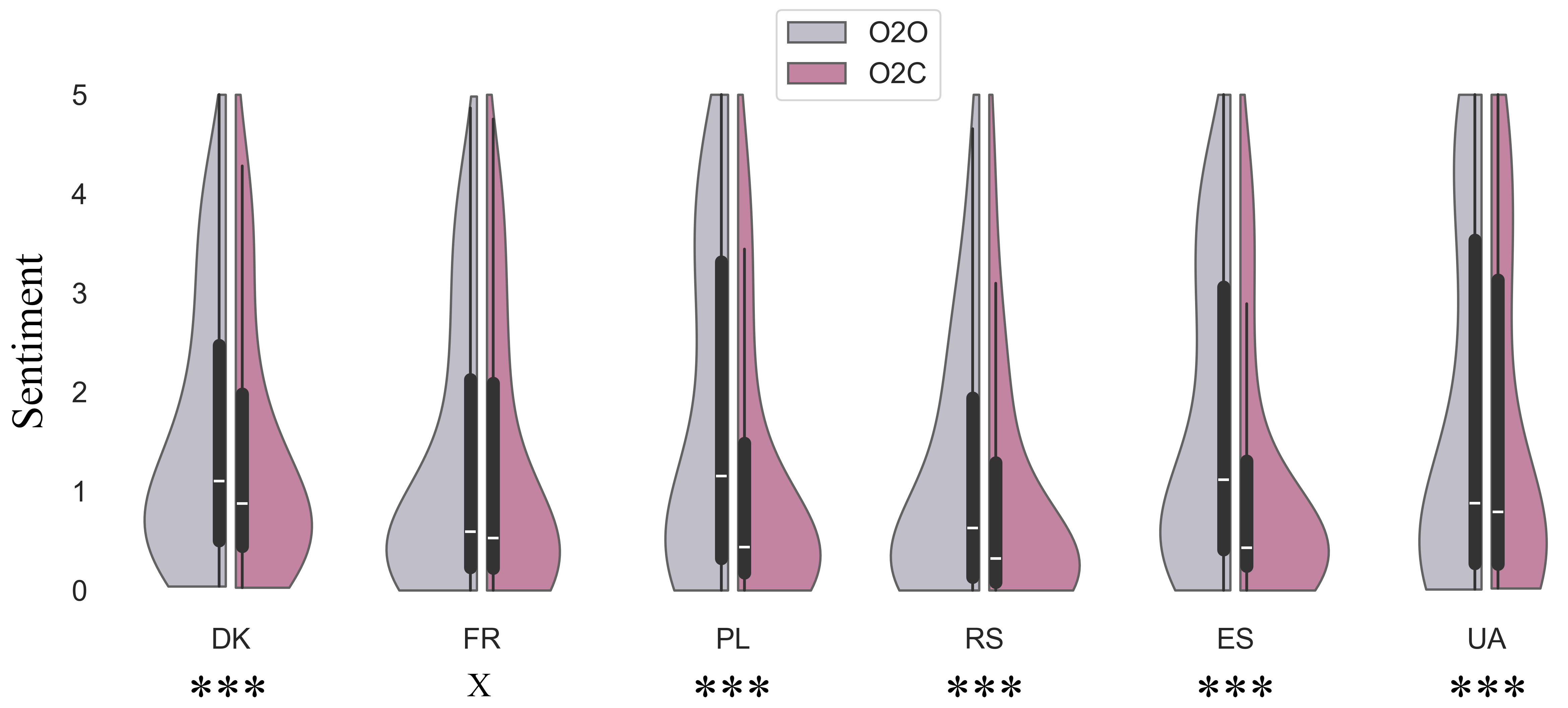}
    \caption{Sentiment distribution of the Opposition towards Opposition (O2O) and towards Coalition (O2C).}
    \vspace{-1em}
    \label{fig:fig3}
\end{figure}

Similarly, we conduct an analogous examination for the Opposition, as illustrated in Figure \ref{fig:fig3}, comparing the O2O sentiment distributions (an Opposition MP referring to an Opposition MP) and the O2C sentiment distributions (an Opposition MP referring to Coalition MP). Despite generally smaller median differences, possibly due to the Opposition's inherent negative stance regardless of the target or topic, results show significant affective polarization of the Opposition for all parliaments, except France.

\subsection{Individual Polarization and Activity}

We investigate whether more frequent debate participation leads to less negativity and reduced affectively polarized behavior by analyzing the relationship between the active participation of individual MPs and their levels of negativity and affective polarization. 

Up to this point, each speech has been treated as an independent entity, without considering the individuality of MPs. To address this, we calculate the sentiment scores for individual MPs and determine the affective polarization by subtracting the sentiment when an MP references the opposing group from the sentiment when referencing their own group.

\begin{figure}[htp!]
    \centering
    \includegraphics[width=1\linewidth]{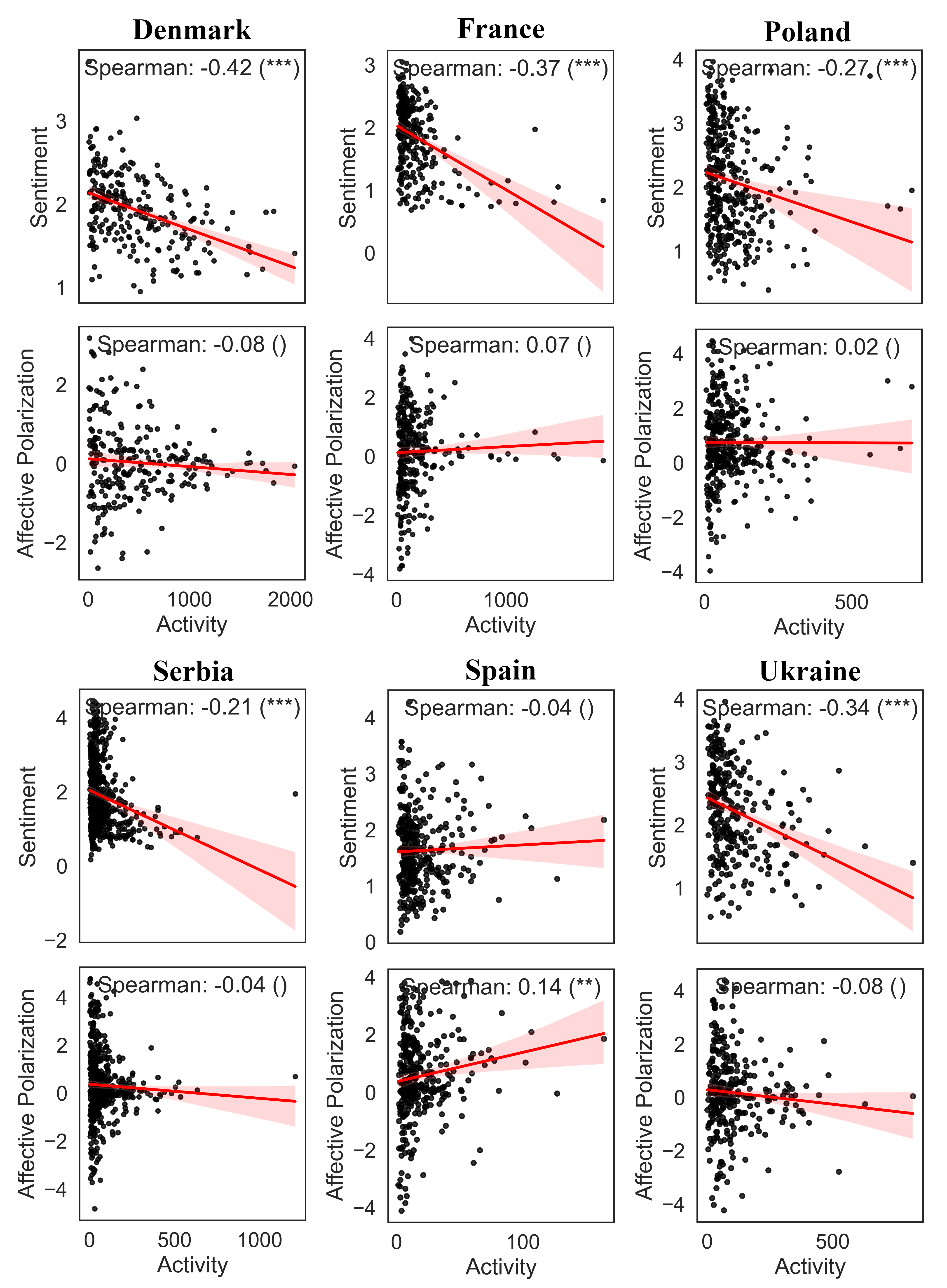}
    \caption{Correlations between activity levels, sentiment, and affective polarization. More active MPs are typically more negative except for Spain where there is no significance.}
    \vspace{-1em}
    \label{fig:fig4}
\end{figure}

Results reveal a weak negative correlation between activity levels of MPs and their sentiment across five out of six parliaments (Spain being the exception), with Spearman's rank correlation values ranging from -0.21 in Serbia to -0.42 in Denmark. This indicates that MPs who deliver more speeches tend to express more negative sentiments. 
On the other hand, we observe no correlation between activity levels of MPs and their affective polarization. These results suggest that, although the less active MPs exhibit lower negativity, they express similar levels of emotional division between Coalition and Opposition (shown in Figure \ref{fig:fig4}).

\subsection{Affective Reciprocity}

Lastly, we investigate whether reciprocity is present in the affective behavior of MPs towards each other, which could help explain overall affective polarization. To assess reciprocity, we compute the Spearman's rank correlation between the sentiment expressed by one MP when referring to another and the sentiment expressed returned by the second MP, across all MP pairs.

\begin{table}[htp!]
\caption{Sentiment reciprocity in the selected parliaments. Values represent the Spearman's rank correlation of speech sentiments in the two directions of a pair of MPs mentioning each other.}
\centering
\begin{tabular}{l|r}
\hline
Parliament & Reciprocity \\ \hline
Denmark (DK)        & 0.10 (***)           \\
France (FR)         & 0.49 (***)           \\
Poland (PL)         & 0.44 (***)           \\
Serbia (RS)         & 0.28 (***)           \\
Spain (ES)          & 0.38 (***)           \\
Ukraine (UA)        & 0.33 (***)          
\end{tabular}
\vspace{-1em}
\label{table:table2}
\end{table}

Results (shown in Table~\ref{table:table2}) suggest that weak to moderate positive reciprocity is present in all six parliaments, with French parliamentarians showing the highest reciprocity, and Danish the lowest. The presence of reciprocity alone does not inherently imply a positive or negative impact on democracy, as it heavily depends upon contextual factors. However, the findings indicate that the nature of parliamentary debate culture is dynamic and varied among parliaments.

\section{Conclusions}

This study introduces a fully automated NLP methodology to investigate affective polarization in transcripts of parliamentary political speech. Results reveal heightened negativity towards opposing groups in all six studied European parliaments, indicating the persistence of affective polarization. Additionally, we found that more active parliamentarians tend to be more negative overall; however, the levels of affective polarization do not correlate with parliamentary activity. We also observed that sentiment reciprocity is present in all six parliaments, potentially serving as a crucial mechanism for establishing affective polarization among parliamentarians.

Our approach involves a notable simplification: using the sentiment expressed in speeches when mentioning a person as a proxy for the sentiment directed towards that person. Depending on the customs and language norms in each parliament, this may not always be accurate. Despite this limitation, the consistency of our results, which significantly exceeds mere noise, suggests the reliability of our method.

This work highlights the usefulness of fully automated methods in analyzing complex social constructs like affective polarization. By leveraging NLP techniques on large-scale parliamentary datasets, we contribute to a deeper understanding of political communication dynamics and the mechanisms underlying affective polarization within political elites. Monitoring democracies through such indicators can provide an early warning mechanism for detecting signs of democratic erosion, helping to maintain democratic resilience.

\noindent \mbox{Code available at:} \href{https://github.com/boevkoski/affective_polarization}{\texttt{github.com/boevkoski/affective\_polarization}}

\begin{table*}[t!]
\footnotesize
\centering
\caption{Statistics of mention detection in speeches of the six parliaments.
Speeches with mentions are those that match exactly one MP in the parliament, excluding self-mentions and speeches mentioning both,
coalition and opposition. Accuracy is the number of correctly identified mentions out of manually checked random samples of 50 speeches for each parliament.
}

\label{tab:ment-stats}
\begin{tabular}{r r r r}
\toprule
\textbf{Parliament} & \textbf{All speeches} & \textbf{Speeches with mentions} (\%) &\textbf{Accuracy} \\
\midrule
Denmark (DK) & 127,049 & 17,732 (13.9\%) & 96\% \\
France (FR) & 57,837 & 5,177 \hphantom{0}(8.9\%) & 100\% \\
Poland (PL) & 59,840 & 8,490 (14.2\%) & 86\% \\
Serbia (RS) & 120,364 & 40,350 (33.5\%) & 92\% \\
Spain (ES) & 16,789 & 4,694 (27.9\%) & 90\% \\
Ukraine (UA) & 60,575 & 5,582 \hphantom{0}(9.2\%) & 96\% \\ 
\bottomrule
\end{tabular}
\label{table:ner_stats}
\end{table*}

\bibliography{custom}

\appendix

\section{Appendix - In-group and out-group member identification}
\label{sec:appendix}

Accurately identifying references in parliamentary speeches is crucial in our methodology for measuring affective polarization. To start, we use ParlaMint's Named Entity Recognition (NER) tags to identify Personal Named Entities (PNEs) for individuals referred to by speakers. We then match PNEs with current MPs by comparing the entire PNE string with full names using token set ratio matching. This strict criterion prevents false positives and ensures each mention is correctly attributed to a single MP. To validate our process, we manually reviewed 50 mentions per parliament. Over 90\% of detected mentions correctly matched intended MPs, affirming our approach's reliability. Table \ref{table:ner_stats} shows results of the reviewing.

\end{document}